\documentclass[conference]{IEEEtran}

\usepackage{graphicx,epsfig,times,amsmath,amsfonts,amssymb,array,enumerate,algorithm, algorithmic} 
\usepackage[subnum]{cases}
\usepackage[latin1]{inputenc}
\usepackage[T1]{fontenc}

\newcommand{\bc}{\mathbf{c}}
\newcommand{\bA}{\mathbf{A}}

\newcommand{\bh}{\mathbf{h}}

\newcommand{\bt}{\mathbf{t}}

\newcommand{\bW}{\mathbf{W}}

\newcommand{\bX}{\mathbf{X}}
\newcommand{\by}{\mathbf{y}}
\newcommand{\bY}{\mathbf{Y}}
\newcommand{\bz}{\mathbf{z}}
\newcommand{\bZ}{\mathbf{Z}}

\newcommand{\bsA}{\boldsymbol{A}}

\newcommand{\cL}{\mathcal{L}}

\newcommand{\cO}{\mathcal{O}}

\newcommand{\bspi}{\boldsymbol{\pi}}

\newcommand{\bsbeta}{\boldsymbol{\beta}}

\newcommand{\bstheta}{\boldsymbol{\theta}}

\newcommand{\bsPsi}{\boldsymbol{\Psi}}

\newcommand{\Identity}{\textbf{I}}

\newcommand{\E}{\mathbb{E}}

\newcommand{\N}{\mathcal{N}}

\IEEEoverridecommandlockouts    

\begin{document}
 
\title{\ \\ \LARGE\bf Model-based clustering with Hidden Markov Model regression for time series with regime changes \thanks{Faicel Chamroukhi is with the Computer Science Lab of Paris Nord University (LIPN, UMR CNRS 7030), Allou Samé and Patrice Aknin are with the Research Unit UPE, IFSTTAR, GRETTIA and Gérard Govaert is with Heudiasyc Lab, UMR CNRS 6599. Contact: Faicel.Chamroukhi@lipn.univ-paris13.fr.} 
}

\author{Faicel Chamroukhi, Allou Sam\'{e}, Patrice Aknin, G\'{e}rard Govaert}
\maketitle

\begin{abstract}
This paper introduces a novel model-based clustering approach for clustering time series which present changes in regime. It consists of a mixture of polynomial regressions governed by hidden Markov chains. The underlying hidden process for each cluster activates successively several polynomial regimes during time. The parameter estimation is performed by the maximum likelihood method through a dedicated Expectation-Maximization (EM) algorithm. The proposed approach is evaluated using simulated time series and real-world time series issued from a railway diagnosis application. Comparisons with existing approaches for  time series clustering, including the stand EM for Gaussian mixtures, $K$-means clustering, the standard mixture of regression models and mixture of Hidden Markov Models, demonstrate the effectiveness of the proposed approach.
\end{abstract}


\section{Introduction}
\label{sec: introduction}
\PARstart   
{T}{he} work presented in this paper relates to the diagnosis of the railway switches which enable trains to be guided from one track to another at a railway junction. The switch is controlled by an electrical motor and the considered time series are the time series of the consumed power during the switch operations. These time series present changes in regime due to successive mechanical motions involved in a switch operation (see Figure \ref{fig. railway times series}). The kind of time series studied here may also  be referred to as longitudinal data, functional data, curves or signals. The diagnosis task can be achieved through the analysis of these time series issued from the switch operations to identify possible faults. However, the large amount of data  makes the manual labeling task onerous for the  experts. Therefore, the main concern of this work is to propose a data preprocessing approach that allows for automatically identifying homogeneous groups in a set of time series. Thus, the founded groups can then be easily treated and interpreted by the maintenance staff in order to identify faults. 
 This preliminary task can be achieved through an unsupervised classification (clustering) approach. In this paper, we focus on model-based clustering approaches for their well established statistical properties and the suitability of the Expectation-Maximization algorithm \cite{dlr} to this unsupervised framework.

In this context, since the time series present regime changes, basic polynomial regression models are not suitable. An alternative approach may consist in using cubic splines to approximate each set of time series \cite{garetjamesJASA2003} but this requires the setting of knots which may a combinatory complex task. Generative models have been developed by Gaffney \& Smyth \cite{Gaffney99trajectoryclustering,gaffneyANDsmythNIPS2004} which consist  in  clustering time series with  mixture of regressions or random effect models. Liu \& Yang \cite{liuANDyangFunctionalDataClustering} proposed a clustering approach based on random effect spline regression where the time series are represented by B-spline basis functions. However, the first approach does not address the problem of changes in regimes and the second one requires the setting of the spline knots. Another approach based on splines is concerned with clustering sparsely sampled time series \cite{garetjamesJASA2003}. We note that all these approaches use the EM algorithm to estimate the model parameters. Another clustering approach consist in the evolutionary clustering approach \cite{Chi_etal_evolutionaryspectral_clustering}, however, in this paper, the structure of the model is fixed over time.

In this paper, a specific generative mixture model is proposed to cluster time series presenting regime changes. In this mixture model, each component density is the one of a specific regression model that incorporates a hidden Markov chain allowing for transitions between different polynomial regression models over time. The proposed model can be seen as an extension of the  model-clustering approach using mixture of standard HMMs introduced by Smyth \cite{Smyth96}, by considering a polynomial regression Hidden Markov Model rather than a standard HMM. In addition, owing to the fact that the real time series of switch operations we aim to model consist of successive phases, order constraints are imposed on the hidden states.

This paper is organized as follows. Section 2 provides an account of the model-based clustering approaches using mixture of regression models and mixture of Hidden Markov Models. Section 3 introduces the proposed model-based time series clustering and its parameter estimation via a dedicated EM algorithm. Finally, section 4 deals with the experimental study carried out on simulated time series and real-world time series of the switch operations to asses the proposed approach by comparing it to existing time series clustering approaches,  in particular, the mixture of regression approach \cite{Gaffney99trajectoryclustering, Gaffneythesis} and the standard mixture of HMMs \cite{Smyth96}.  

\section{Model-based clustering for time series}
\subsection{Model-based clustering}
\label{sec: Model-based clustering}
Model-based clustering \cite{banfield_and_raftery_93,mclachlan_basford88,Fraley2002_model-basedclustering}, generally used for multidimensional data, is based on the finite mixture model formulation \cite{mclachlanFiniteMixtureModels}. 
 In the finite mixture approach for cluster analysis, the data probability density function is assumed to be a mixture of $K$ components densities, each component density being associated with a cluster. The problem of clustering therefore becomes the one of estimating the parameters of the assumed mixture model (e.g, estimating the mean vectors and the covariance matrices in the case of Gaussian mixtures). The parameters of the mixture density are generally estimated by maximizing the observed-data likelihood via the well-known Expectation-Maximization (EM) algorithm \cite{dlr,mclachlanEM}. After performing the probability density estimation, the obtained posterior cluster probabilities are then used to determine the cluster memberships through the maximum a posteriori (MAP) principle and therefore to provide a partition of the data into $K$ clusters.

Model-based clustering approaches have also been introduced to generalize the standard multivariate mixture model for the analysis of time series data, which are also referred to as longitudinal data, functional data or sequences. In that case, the individuals are presented as functions or curves rather than a vector of a reduced dimension. 
In that context, one can distinguish the  regression mixture approaches \cite{Gaffney99trajectoryclustering, Gaffneythesis}, including polynomial regression and spline regression. Random effects approaches that are based on polynomial regression \cite{gaffneyANDsmythNIPS2004} or spline regression \cite{liuANDyangFunctionalDataClustering}.  Another approach based on splines is concerned with clustering sparsely sampled time series \cite{garetjamesJASA2003}. All these approaches use the EM algorithm to estimate the model parameters. 
In the following section we will give an overview of these model-based clustering approaches for time series.

Let $\bY=(\by_1,\ldots,\by_n)$ be a set of $n$  independent time series and let $(h_1,\ldots,h_n)$ be the associated unknown cluster labels  with $h_i \in \{1,\ldots,K\}$. We assume that each time series $\by_i$ consists of $m$  measurements (or observations) $\by_i = (y_{i1},\ldots,y_{im})$, regularly observed at the time points $\bt =(t_1,\ldots,t_m)$ with $t_1<\ldots<t_m$.
  
\subsection{Related work on model-based clustering for time series}
\label{sec: related work for time series clustering}
 
\subsubsection{Mixture of regression models} 
\label{ssec: polynomial and spline regression mixture}

In this section we describe  time series clustering approaches based on polynomial regression mixtures  and polynomial spline regression mixtures \cite{Gaffney99trajectoryclustering, Gaffneythesis}. The regression mixture approaches assume that each times series is drawn from one of $K$ clusters of time series which are mixed at random in proportion to the relative cluster sizes $(\alpha_1,\ldots,\alpha_K)$. Each cluster of time series is modeled by either a  polynomial regression model or a spline regression model. Thus, the conditional mixture density of a time series $\by_i$ 
can be written as: 
\begin{IEEEeqnarray}{lcl}
f(\by_i|\bt;\bsPsi)&=& \sum_{k=1}^K \alpha_k \  \N (\by_{i};\bX \bsbeta_k,\sigma_k^2\Identity_m), 
\label{eq: PRM or PSRM model}
\end{IEEEeqnarray}
where the $\alpha_k$'s defined by $\alpha_k = p(h_i=k)$ are the non-negative mixing proportions that sum to 1, $\bsbeta_{k}$ is the $(p+1)$-dimensional coefficient  vector of the $k$th polynomial regression model, $p$ being the polynomial degree, and $\sigma_{k}^2$ is the associated noise variance. The matrix $\bX$ is the $m\times(p+1)$ design  matrix with rows $\bt_j = (1,t_j,t_j^2,\ldots,t_j^p)$ for $j=1,\ldots,m$ and $\Identity_m$ is the identity matrix of dimension $m$. The  model  is therefore described by the parameter vector
 $\bsPsi = (\alpha_1,\ldots,\alpha_k,\bsPsi_1,\ldots,\bsPsi_K)$ with  $\bsPsi_k=(\bsbeta_{k},\sigma_{k}^2)$.

Parameter estimation is performed by maximizing the observed-data log-likelihood of $\bsPsi$:
\begin{IEEEeqnarray}{lcl}
\cL(\bsPsi) &=& \sum_{i=1}^n  \log \sum_{k=1}^K \alpha_k \ \N (\by_{i};\bX \bsbeta_k,\sigma_k^2\Identity_m).
\label{eq: log-lik for the PRM and PSRM}
\end{IEEEeqnarray}
This log-likelihood, which can not be maximized in a closed form, is maximized by the EM algorithm \cite{dlr}. The details of the EM algorithm for the mixture of regressions models and the corresponding updating formula can be found in \cite{Gaffney99trajectoryclustering, Gaffneythesis}.

Once the model parameters are estimated, a partition of the data is then computed by maximizing the posterior cluster probabilities defined by:
\begin{IEEEeqnarray}{lcl}
\!\!\!\!\!\! \tau_{ik} &  =  & p(h_i=k|\by_{i},\bt;\bsPsi) 
\!\! = \!\! \frac{\alpha_k \N\big(\by_i;\bX \bsbeta^{T}_{k},\sigma^{2}_{k}\Identity_m\big)}{\sum_{k'=1}^R \alpha_{k'} \N(\by_i;\bX \bsbeta^{T}_{k'},\sigma^{2}_{k'}\Identity_m)}\cdot
\label{eq: post prob h_ir of the cluster r for the mixture of piecewise regression}
\end{IEEEeqnarray}

The mixture of regression models however  do not address the problem of regime changes within times series. Indeed, they assume that each cluster present a stationary behavior described by a single polynomial mean function. The spline regression mixture 
does not address automatically the regime changes as the knots are generally fixed in advance and the optimization of their location needs a strong computational load. These approach may therefore have limitations in the case of time series presenting changes in regime. To overcome these limitations, one way is to proceed as in the case of sequential data modeling in which it is assumed that the observed sequence (in this case a times series) is governed by a hidden process which enables for switching from one state to another among $R$ states. The used process in general is an $R$ state Markov chain for each time series. This leads to the mixture of Hidden Markov Models \cite{Smyth96} which we describe in the following section.

\subsubsection{Mixture of HMMs for clustering sequences} 
\label{ssec: clustering with HMMs}
In this section we describe the mixture of Hidden Markov Models (HMMs) initiated by Smyth \cite{Smyth96} and used for clustering sequences, which can therefore be applied to time series. Since the model in this case includes an HMM formulation, let us first recall the principle of HMMs.
 
\paragraph{Hidden Markov Models (HMMs)}

Hidden Markov Models (HMMs) are a class of latent data models appropriate for sequential data. They are widely used in many application domains, including speech recognition, image analysis, time series prediction \cite{rabiner, Derrode2006}, etc.  In an HMM, the observation sequence (or a time series) $\by_i=(y_{i1},\ldots,y_{im})$ is assumed to be governed by a hidden state sequence $\bz_i = (z_{i1},\ldots,z_{im})$ where the discrete random variable $z_{ij} \in \{1,\ldots,R\}$  represents the unobserved state associated with $y_{ij}$ at instant $t_j$.  The  state sequence $\bz_i$ is generally assumed to be a first order homogeneous Markov chain, that is, the current state given the previous state sequence depends only on the previous state. Formally we have :
\begin{equation}
p(z_{i1} | z_{i,j-1}, z_{i,j-2}, \ldots, z_{i1}) = p(z_{ij}|z_{i,j-1}) \ \forall j>1.
\label{eq: Markov property}
\end{equation}
The transition probabilities $p(z_{ij}|z_{i,j-1})$ do not depend on $t$ in the case of  an homogeneous Markov chain. An  HMM is therefore fully determined by 
the initial state distribution $\bspi =(\pi_1,\ldots,\pi_R)$ where $\pi_r = p(z_1 = r)$ satisfying $\sum_{r}  \pi_r=1$, 
the matrix of transition probabilities  $\bsA$ with elements $A_{\ell r} = p(z_{ij}=r|z_{i,j-1}=\ell)$ satisfying $\sum_{r}  A_{\ell r }=1$ and
the parameters $(\bsPsi_1,\ldots,\bsPsi_R)$ of the { emission probabilities} $p(y_{ij}|z_{ij}=r;\bsPsi_r)$. 
The distribution of a particular configuration  of the latent state sequence $\bz_i=(z_{i1},\ldots,z_{im})$ is is given by:
\begin{equation}
p(\bz_i;\bspi,\bsA) = p(z_{i1};\bspi) \prod_{j=2}^{m}p(z_{ij}|z_{i,j-1};\bsA),
\label{eq: general HMM_process (Markov chain)}
\end{equation}
and from the conditional independence property of the HMM, that is the observation sequence is independent given a particular configuration of the hidden state sequence, the conditional distribution of the observed sequence is therefore given by:

\begin{equation}
p(\by_i|\bz_i;\bsPsi) =\prod_{j=1}^{m}p(y_{ij}|z_{ij};\bsPsi).
\label{eq: cond. density of the observed sequence for an HMM}
\end{equation}
From (\ref{eq: general HMM_process (Markov chain)}) and (\ref{eq: cond. density of the observed sequence for an HMM}), we can then get  the following joint distribution  $p(\by_i,\bz_i;\bsPsi) = p(\bz_i;\bspi,\bsA)p(\by_i|\bz_i;\bsPsi).$ 

\paragraph{Mixture of Hidden Markov Models}

The mixture of HMMs integrates the HMM into a mixture framework to perform sequence clustering \cite{Smyth96, Alon_03}. In this probabilistic model-based clustering, an observation sequence (in this case a time series) is assumed to be generated according to a mixture of $K$ components, each component being an HMM. Formally, each time series $\by_i$ is distributed according to the following mixture distribution:
 \begin{equation}
f(\by_i;\bsPsi) = \sum_{k=1}^K \alpha_k f_k(\by_i;\bsPsi_k), 
\label{eq. MixHMM model}
\end{equation}
where  the component density $f_k(\by_i;\bsPsi_k) = p(\by_i|h_i=k;\bsPsi_k)$ is assumed to be a $K$ state HMM, typically with univariate Gaussian emission probabilities in this case of univariate time series. The HMM associated with the $k$th cluster is determined by the parameters $\bsPsi_k=(\bspi_k, \bsA_k,\mu_{k1},\ldots,\mu_{kR},\sigma^2_{k1},\ldots,\sigma^2_{kR})$ where $\bspi_k$ is the initial state distribution for the HMM associated with cluster $k$, $\bsA_k$ is the corresponding transition matrix and $(\mu_{kr},\sigma^2_{kr})$ are respectively the constant mean and the variance of an univariate Gaussian density associated with the $r$th state in cluster $k$. By using the joint distribution of $\by$ and $\bz$ which can be deduced from (\ref{eq: general HMM_process (Markov chain)}) and (\ref{eq: cond. density of the observed sequence for an HMM}),  the distribution of a time series issued from the $k$th cluster is therefore given by: 
\begin{IEEEeqnarray}{lcl}
f_k(\by_i;\bsPsi_k) &=&   \sum_{\bz_i}  p(z_{i1};\bspi_k) \prod_{j=2}^{m}p(z_{ij}|z_{i,j-1};\bsA_k)\times \IEEEnonumber \\
&&\prod_{j=1}^{m} \N(y_{ij};\mu_{kz_{ij}},\sigma_{kz_{ij}}^2). 
\label{eq: conditional density for a Gaussian hmm of first order}
\end{IEEEeqnarray}

Two different approaches can be adopted for estimating this mixture of HMMs. Two such  techniques are the hard-clustering $K$-means-like approach and the soft-clustering EM approach. The $K$-means-like approach for hard clustering have been used in \cite{Smyth96} in which the optimized function is the complete-data log-likelihood. The resulting clustering scheme consists of assigning sequences to clusters at each iteration and using only the sequences assigned to a cluster for re-estimation of its HMM parameters.  The soft clustering approach  is described in \cite{Alon_03} where the model parameters are estimated in a maximum likelihood framework by the EM algorithm. 

In this standard mixture of HMMs, each state is represented by its scalar mean in the case of univariate time series.  However, in many applications,  in particular in signal processing or time series analysis, as in the case of the time series issued from the switch operations, it is often useful to represent a state by a polynomial rather than a scalar (constant function of time). This assumption should be more suitable for fitting the non-linear regimes governing the time series. In addition, when the regimes are ordered in time, the hidden process governing the time series can be adapted by imposing order constraints on the states of the Markov chain. These generalizations are integrated in the proposed mixture of HMM regression models which we present in the following section.

\section{The proposed mixture of HMM regression models for time series clustering}


\subsection{Model definition}
\begin{figure}
\centering
\includegraphics[scale=.15]{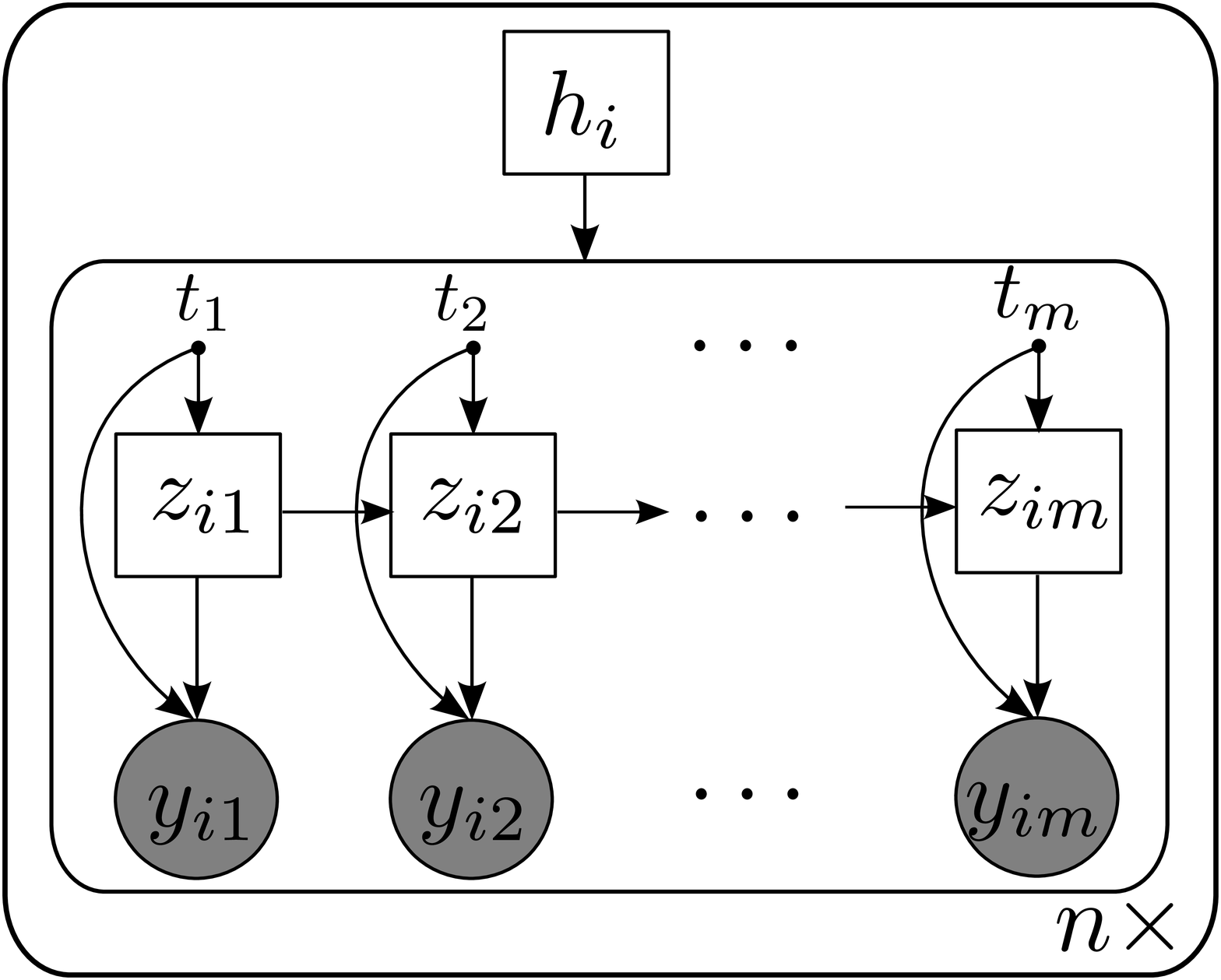} 
\caption{Graphical model structure for the proposed mixture of HMM regression models (MixHMMR).}
\label{fig: graphical model for the MixHMMR}
\end{figure}

The proposed model assumes that each time series $\by_i$ is issued from one of $K$ clusters where, within each cluster $k$ $(k=1,\ldots,K)$, each time series is generated by $R$ unobserved polynomial regimes. The transition from one regime to another is governed by an homogeneous Markov Chain of first order. 
Formally, the distribution of a times series $\by_i$ is defined by the following conditional  mixture density:
\begin{equation}
f(\by_i |\bt;\bsPsi) =\sum_{k=1}^K \alpha_k f_k(\by_i|\bt;\bsPsi_k),
\label{eq. MixHMMR model}
\end{equation}
where each component density $f_k(.)$ associated with the $k$th cluster is a polynomial HMM regression model (see \cite{fridman} for details on HMM regression for a single time series). In this clustering context with HMM regression, given the cluster $h_i=k$, the time series $\by_i=(y_{i1},\ldots,y_{im})$ is assumed to be generated by the following regression model :
\begin{equation}
y_{ij} = \bsbeta^T_{kz_{ij}}\bt_{j} + \sigma_{k z_{ij}}\epsilon_{ij} \quad (j=1,\ldots,m)
\label{eq: HMM regression model}
\end{equation}
where $\bsbeta_{kr}$ is the $(p+1)$-dimensional coefficients vector of the $r$th polynomial regression model of  cluster $k$, $\sigma_{kr}^2$ is its associated noise variance and the $\epsilon_{ij}$ are independent random variables distributed according to a Gaussian distribution with zero mean and unit variance.  The hidden state sequence $\bz_i=(z_{i1},\ldots,z_{im})$ is assumed to be 
Markov chain  of parameters $(\bspi_k,\bA_k)$.
The proposed model is illustrated by the graphical representation in Figure \ref{fig: graphical model for the MixHMMR}. 
Each component density is therefore parametrized by the parameter vector
$\bsPsi_k= (\bspi_k,\bsA_k,\bsbeta_{k1},\ldots,\bsbeta_{kR},\sigma_{k1}^2,\ldots,\sigma_{kR}^2)$ and is given in a similar way as for (\ref{eq: conditional density for a Gaussian hmm of first order}) by: 
\begin{IEEEeqnarray}{lCl}
f_k(\by_i|\bt;\bsPsi_k)  &=& \sum_{\bz_i} p(z_{i1};\bspi_k)\prod_{j=2}^{m}p(z_{ij}|z_{i,j-1};\bsA_k) \times \nonumber \\
& & \prod_{j=1}^{m}\N(y_{ij};\bsbeta^T_{kz_{ij}}\bt_j,\sigma_{kz_{ij}}^2).
\label{eq. conditional time series distribution for the MixHMMR (2)}
\end{IEEEeqnarray} 

\subsection{A HMMR with order constraints}
\label{constraint transmat}

Since the time series we aim to model here consist of successive contiguous regimes, we impose order constraints on the hidden states by imposing the following constraints on the transition probabilities for each cluster $k$.  These constraints imply that no transitions are allowed for the phases whose indexes are lower than the current phase and no jumps of more than one state are possible. Formally, we have:
$$A_{k\ell r}=p(z_{ijk}=r|z_{i(j-1)k}= \ell, h_i =k) = 0  \ \mbox{if} \ r<\ell$$
and
$$A_{k\ell r}=p(z_{ijk}=r|z_{i(j-1)k}= \ell, h_i =k) = 0 \ \mbox{if} \  r>\ell+ 1.$$
This constrained model is a  particular case of the well known left-right model \cite{rabiner}.

\subsection{Remark: Link with the polynomial regression  mixture}
 
The particular case for which the proposed model is defined with a single regime $R=1$ for each cluster $k$, corresponds to the polynomial regression mixture model. 

The next section presents the parameter estimation by the maximum likelihood method.

\subsection{Parameter estimation}
\label{sec: parameter estimation}
 
The proposed MixHMMR model is described by the parameter vector
$\bsPsi = (\alpha_1,\ldots,\alpha_K, \bsPsi_1,\ldots,\bsPsi_K).$ Parameter estimation is performed by maximizing the observed-data log-likelihood of $\bsPsi$ :
\begin{IEEEeqnarray}{lll}
\cL(\bsPsi)&=&\log p(\by_1,\ldots,\by_n|\bt;\bsPsi) = \log  \prod_{i=1}^n p(\by_i|\bt;\bsPsi)\IEEEnonumber\\
&=& \sum_{i=1}^n \log \sum_{k=1}^K \alpha_k \sum_{\bz_i} p(z_{i1};\bspi_k)\prod_{j=2}^{m}p(z_{ij}|z_{i,j-1};\bsA_k) \times \IEEEnonumber \\
& & \prod_{j=1}^{m}\N(y_{ij};\bsbeta^T_{kz_{ij}}\bt_j,\sigma_{kz_{ij}}^2).
\label{eq. conditional time series distribution for the MixHMMR (2)}
\label{eq: log-likelihood MixHMMR model} 
\end{IEEEeqnarray} 
The maximization of this log-likelihood cannot be performed in a closed form. We maximize it iteratively by using a dedicated EM algorithm.  With this specification   of the EM algorithm, the complete-data for the proposed model consist of the observed set of curves $\bY=(\by_1,\ldots,\by_n)$, their corresponding cluster labels $\bh=(h_1,\ldots,h_n)$ and the matrix of regime (state) labels $\bZ = (\bz_1,\ldots,\bz_n)$, $\bz_i$ being the hidden state sequence associated with $\by_i$. The complete-data likelihood of $\bsPsi$ is therefore given by:
{  \begin{IEEEeqnarray*}{lll}
 p(\bY,\bh,\bZ|\bt;\bsPsi) &=& p(\bh)p(\bY,\bZ|\bh,\bt;\bsPsi) \nonumber\\
&=& p(\bh)p(\bZ|\bh,\bt;\bsPsi) p(\bY|\bh,\bZ,\bt;\bsPsi) \nonumber\\
&=&  \prod_{i=1}^n p(h_i) p(\bz_i|\bt;\bspi_{h_i},\bsA_{h_i}) p(\by_i|\bz_i,\bt;\bstheta_{h_i}).  \nonumber\\
\label{eq: first expression of complete-data likelihood for the MixHMMR}
\end{IEEEeqnarray*}}
Then, by using some elementary calculation details, we get the complete complete-data log-likelihood: 
{\begin{IEEEeqnarray}{lll}
\!\!\!\!\!\!\!\!\! \cL_c(\bsPsi) 
&=& \log  p(\bY,\bh,\bZ|\bt;\bsPsi) \IEEEnonumber \\
%
&=& \sum_{k=1}^K  \Big[ \sum_{i=1}^{n} h_{ik} \log \alpha_k  +   \sum_{i=1}^{n}  \sum_{r=1}^{R} h_{ik} z_{i1kr} \log \pi_{kr}  \IEEEnonumber \\
&& + \sum_{i=1}^{n} \sum_{j=2}^{m} \sum_{r,\ell=1}^{R} h_{ir} z_{ijkr} z_{i(j-1)k \ell} \log A_{k \ell r}  \IEEEnonumber \\
&& + \sum_{i=1}^{n} \sum_{j=1}^{m}  \sum_{r=1}^{R} h_{ik} z_{ijkr}\log \mathcal{N} (y_{ij};\bsbeta^T_{kr}\bt_{j},\sigma^2_{kr}) \Big]
\end{IEEEeqnarray}}
where we have used the following indicator binary variables for indicating the cluster memberships and the regime meberships for a given cluster, that is:
\begin{itemize}
\item  $h_{ik}=1$ if $h_i=k$ (i.e., $\by_i$ belongs to cluster $k$) and $h_{ik}=0$ otherwise.
\item  $z_{ijkr}=1$ if $z_{ijk}=r$ (i.e., the $i$th times series $\by_i$ belongs to cluster $k$ and its $m$th observation $y_{ij}$ belongs to regime $r$) and $z_{ijkr}=0$ otherwise.
\end{itemize}

The next section gives the proposed EM algorithm for the mixture of HMM regression models.

\subsubsection{The dedicated EM algorithm}
\label{ssec. EM algortihm}

The   EM algorithm for the proposed MixHMMR model starts from an initial parameter $\bsPsi^{(0)}$ and alternates between the two following steps until convergence:
\paragraph{ E Step} Compute the expected complete-data log-likelihood  given the time series $\bY$, the time vector $\bt$ and the current value of the parameter $\bsPsi$ denoted by  $\bsPsi^{(q)}$:  
 \begin{IEEEeqnarray}{lcl}
Q(\bsPsi,\bsPsi^{(q)})&  =   & \E\big[\cL_c(\bsPsi)|\bY,\bt;\bsPsi^{(q)}\big]  
\label{eq: Q-function for the MixHMMR (1)}
\end{IEEEeqnarray} 
It can be easily shown that this conditional expectation   is given by:   
{  \begin{equation} 
 Q(\bsPsi,\bsPsi^{(q)}) = Q_1(\alpha_k)   + \sum_{k=1}^{K} \Big[Q_2(\bspi_k,\bA_k)  + Q_3(\bsbeta_{kr},\sigma^2_{kr})\Big],\nonumber \\
\label{eq: Q-function for the MixHMMR model}
\end{equation}}
where 
{\small $$Q_1(\alpha_k) = \sum_{k=1}^K \sum_{i=1}^{n} \tau_{ik}^{(q)} \log \alpha_k,$$}  
{\small $$Q_2(\bspi_k,\bsA_k) =\sum_{r=1}^{R} \sum_{i=1}^{n} \tau_{ik}^{(q)} \big[\gamma^{(q)}_{ik1} \log \pi_{kr} + \sum_{j=2}^{m} \sum_{\ell=1}^{R} \xi^{(q)}_{ijk\ell r} \log A_{k \ell r}\big],$$}
{\small $$Q_3(\bsbeta_{kr},\sigma^2_{kr}) =\sum_{r=1}^R \sum_{i=1}^{n} \sum_{j=1}^{m} \tau_{ik}^{(q)} \gamma^{(q)}_{ijkr}\log \mathcal{N} (y_{ij};\bsbeta^T_{kr}\bt_{j},\sigma^2_{kr})$$}
where
\begin{itemize}
\item $\tau^{(q)}_{ik}  =p(h_i=k|\by_i,\bt;\bsPsi^{(q)})$ is the posterior probability of  cluster $k$;
\item $\gamma^{(q)}_{ijkr} = p(z_{ijk}=r|\by_i,\bt;\bsPsi_k^{(q)})$ is the posterior probability of the $k$th polynomial regime for the $k$th cluster,
\item $\xi^{(q)}_{ijkr\ell}  = p(z_{ijk}=r, z_{i(j-1)k}=\ell|\by_i,\bt;\bsPsi_k^{(q)})$ is the joint probability of having the regime $r$ at time $t_j$ and the regime $\ell$ at time $t_{j-1}$ in cluster $k$.
\end{itemize}
As shown in the expression of $Q$, this step requires only the computation of the probabilities $\tau^{(q)}_{ik}$, $\gamma^{(q)}_{ijkr}$ and $\xi^{(q)}_{ijk\ell r}$. The probabilities $\gamma^{(q)}_{ijkr}$ and $\xi^{(q)}_{ij k\ell r}$ for each time series $\by_i$ $(i=1,\ldots,n)$ are computed as follows \cite{rabiner}:
\begin{equation}
\gamma^{(q)}_{ijkr} =\frac{a^{(q)}_{ijkr} b^{(q)}_{ijkr}}{\sum_{\ell = 1}^R a^{(q)}_{ij k \ell} b^{(q)}_{ijk\ell}}
\label{eq. regime post prob for an HMMR}
\end{equation}
and
\begin{equation}
 \xi^{(q)}_{ij k \ell r}=\frac{a^{(q)}_{i(j-1)\ell} \ A^{(q)}_{k\ell r} \N(y_{ij};\bsbeta^{(q)T}_{kr}\bt_j,\sigma_{kr}^{(q)2}) b^{(q)}_{ijkr}}{\sum_{r,\ell=1}^R a^{(q)}_{i(j-1)\ell k} \ A^{(q)}_{k\ell r} \N(y_{ij};\bsbeta^{(q)T}_{kr}\bt_j,\sigma_{kr}^{(q)2}) \ b^{(q)}_{ijkr}}.
\label{eq. joint regime post prob for an HMMR}
\end{equation}
where the quantities  $a_{ijkr}$ and $b_{ijkr}$ are respectively the forward probabilities   and the backward probabilities, which are in this context given by:
\begin{equation}
a_{ijkr} = p(y_{i1},\ldots, y_{ij},z_{ijk}=r|\bt;\bsPsi_k),
\label{eq: forward variables for a HMM regression}
\end{equation} 
and
\begin{equation}
b_{ijkr} = p(y_{i,j+1},\ldots,y_{im}|z_{ijk}=r,|\bt;\bsPsi_k) 
\label{eq: backward variables for a HMM regression}
\end{equation}
and are recursively computed via the well-known forward-backward (Baum-Welch) procedure \cite{BaumWelch, rabiner}.
  
The posterior cluster probabilities $\tau_{ik}^{(q)}$ that the time series $\by_i$ belongs to cluster $k$ are computed as follows:
\begin{equation}
\tau^{(q)}_{ik} = \frac{\alpha_k^{(q)} f_k(\by_i|\bt;\bsPsi^{(q)}_k)}{\sum_{k'=1}^K \alpha_{k'}^{(q)}f_{k'}(\by_i|\bt;\bsPsi^{(q)}_{k'})},
\label{eq: cluster post prob for the MixHMMR}
\end{equation}
where the conditional probability distribution of the time series $\by_i$ given a cluster $k$, which can be expressed in function of the forward variables $a_{ijkr}$ (\ref{eq: forward variables for a HMM regression}) as:
$$f_k(\by_i|\bt;\bsPsi^{(q)}_k)=p(y_{i1},\ldots,y_{im}|\bt;\bsPsi^{(q)}_k) = \sum_{r=1}^{R} a_{imkr},$$
is therefore obtained after the forward procedure.

\paragraph{M-step}
\label{par: M-step of the EM algorithm for the MixHMMR model}

In this step, the value of the parameter $\bsPsi$ is updated by  maximizing the expected complete-data log-likelihood  with respect to $\bsPsi$, that is:
\begin{equation}
\bsPsi^{(q+1)} = \arg \max_{\bsPsi} Q(\bsPsi,\bsPsi^{(q)}). 
\end{equation} 

The maximization of $Q$ can be performed by separately maximizing the functions $Q_1$, $Q_2$ and $Q_3$.  
The maximization of $Q_1$ w.r.t the mixing proportions $\alpha_k$ is the one of a standard mixture model. The updates are given by:
\begin{equation}
{\alpha_k}^{(q+1)} = \frac{\sum_{i=1}^n \tau_{ik}^{(q)}}{n}\cdot
\label{eq. mixing prop update for the MixHMMR}
\end{equation}
The maximization of $Q_2$ w.r.t  the parameters $(\bspi_k, \bA_k)$ correspond to a weighted version of updating the parameters of the Markov chain in a standard HMM. The weights in this case are the posterior cluster probabilities $\tau_{ik}$ and the updates are given by:
\begin{equation}
\pi_{kr}^{(q+1)} = \frac{\sum_{i=1}^{n} \tau_{ik}^{(q)} \gamma^{(q)}_{i1kr}}{\sum_{i=1}^{n}  \tau_{ik}^{(q)}},
\label{eq. initial state prob update for the MixHMMR}
\end{equation}
and 
\begin{equation}
A^{(q+1)}_{k\ell r})= \frac{\sum_{i=1}^{n} \sum_{j=2}^{m} \tau_{ik}^{(q)} \xi^{(q)}_{ijk\ell r}}{\sum_{i=1}^{n} \sum_{j=2}^{m} \tau_{ik}^{(q)} \gamma^{(q)}_{ijkr}}\cdot
\label{eq. trans mat update for the MixHMMR}
\end{equation}
Maximizing  $Q_3$ with respect to regression parameters $\bsbeta_{kr}$ for $k=1,\ldots,K$ and $r=1,\ldots,R$ consists in analytically solving $K \times R$  weighted least-squares problems where the weights consists in both the posterior cluster probabilities $\tau_{ik}$ and the posterior regimes probabilities $\gamma^{(q)}_{ijkr}$ for each cluster $k$. The parameter updates are given by:
\begin{equation}
{\bsbeta}_{kr}^{(q+1)}  
=\Big[\bX^T \big(\sum_{i=1}^{n} \tau_{ik}^{(q)} \bW_{ikr}^{(q)}\big) \bX \Big]^{-1} \bX^T \big(\sum_{i=1}^{n} \tau_{ik}^{(q)} \bW_{ikr}^{(q)} \by_i \big),
\label{eq: regression param update for the MixHMMR}
\end{equation}
where $\bW_{ikr}^{(q)}$ is an $m$ by $m$ diagonal matrix whose diagonal elements are the weights $\{\gamma_{ijkr}^{(q)}; j=1,\ldots,m\}$.

Finally, the maximization of $Q_3$ with respect to noise variances $\sigma_{kr}^{2(q+1)}$ consists in a weighted variant of the problem of estimating the variance of an univariate  Gaussian density.  The updating formula is given by: 
\begin{equation}
\sigma_{kr}^{2(q+1)}= 
  \frac{\sum_{i=1}^{n}  \tau_{ik}^{(q)} |\!| \sqrt{\bW_{ikr}^{(q)}} (\by_i -  \bX \bsbeta_{kr}^{(q+1)}) |\!|^2}{ \sum_{i=1}^n \tau_{ik}^{(q)} \text{trace}(\bW_{irk}^{(q)}) },
\label{eq: variance update for the MixHMMR}
\end{equation}
where $|\!|\cdot|\!|$ is the euclidian norm.

The pseudo code \ref{algo: EM algorithm for the MixHMMR model} summarizes the EM algorithm for the proposed MixHMMR model.
\begin{algorithm}
\caption{\label{algo: EM algorithm for the MixHMMR model} Pseudo code of the proposed algorithm.}
{\bf Inputs:}  \hspace*{-.4cm}$(\by_1,\ldots,\by_n), (t_1,\ldots,t_m), K, R, p$
\begin{algorithmic}[1]
\STATE \textbf{Initialize:} $\bsPsi^{(0)}= (\alpha^{(0)}_1,\ldots,\alpha^{(0)}_R,\bsPsi_1^{(0)},\ldots,\bsPsi_K^{(0)})$
\STATE fix a threshold $\epsilon>0$
\STATE set $q \leftarrow 0$ (EM iteration)
\WHILE {increment in log-likelihood $> \epsilon$}
\STATE \underline{E-Step}: 
	   \FOR{$k=1,\ldots,K$}	
	   \STATE  forward-backward procedure:	
			\FOR{$r=1,\ldots,R$}
				\STATE  compute $\gamma_{ijkr}^{(q)}$ for $i=1,\ldots,n$ and $j=1,\ldots,m$ using Equation (\ref{eq. regime post prob for an HMMR})
							\FOR{$\ell=1,\ldots,R$}
								\STATE compute $\xi_{ijk\ell r}^{(q)}$ for $i=1,\ldots,n$ and $j=1,\ldots,m$ using Equation (\ref{eq. joint regime post prob for an HMMR})
			\ENDFOR
			\ENDFOR
				   	\STATE compute $\tau_{ik}^{(q)}$ for $i=1,\ldots,n$ using Equation (\ref{eq: cluster post prob for the MixHMMR}) 
	 \ENDFOR
\STATE \underline{M-Step}:
	\FOR{$k=1,\ldots,K$}	
		   	\STATE compute 
		   	$\alpha_{k}^{(q+1)}$ using Equation (\ref{eq. mixing prop update for the MixHMMR})	
		\FOR{for $r=1,\ldots,R$}
			\STATE compute $\bspi_{kr}^{(q+1)}$ using Equation (\ref{eq. initial state prob update for the MixHMMR})
			\STATE compute $\bsA_{k\ell r}^{(q+1)}$ using Equation (\ref{eq. trans mat update for the MixHMMR})
			\STATE compute $\bsbeta_{kr}^{(q+1)}$ using Equation (\ref{eq: regression param update for the MixHMMR})
			\STATE compute $\sigma_{kr}^{2(q+1)}$ using Equation (\ref{eq: variance update for the MixHMMR})
		\ENDFOR
	\STATE $q \leftarrow q+1$
	  \ENDFOR 
	\ENDWHILE
\STATE $\hat{\bsPsi}= (\alpha^{(q)}_1,\ldots,\alpha^{(q)}_R, \bsPsi^{(q)}_1,\ldots \bsPsi^{(q)}_R)$ 
\end{algorithmic}
\end{algorithm}

\subsubsection{Model selection}
\label{sec: model selection mixture of rhlp}

The problem of model selection is the one of estimating the optimal values of the number of clusters $K$, the number of regimes $R$ and the polynomial degree $p$. The best values $(K,R,p)$ can be computed by maximizing the BIC criterion \cite{BIC} defined by:
\begin{equation}
\mbox{BIC}(K,R,p)=\cL(\hat{\bsPsi})-\frac{\nu(K,R,p)}{2} \log(n),
\label{eq: BIC(R) for FMLDA}
\end{equation}
where $\hat{\bsPsi}$  is the maximum likelihood estimate of the parameter vector $\bsPsi$ provided by the EM algorithm, $\nu(K,R,p)= K-1 + KR + K(R+R-1) + KR(p+1) + KR$ is the number of free parameters of the MixHMMR model which is  respectively  composed of the free mixing proportions ($K-1$), the number of initial state probabilities ($KR$), the number of free transitions probabilities ($K(R+R-1$), the number of regression coefficients ($KR(p+1)$) and the number of variances ($KR$), $n$ being the sample size. 
The BIC values are computed for $K$  varying  from $1$ to $K_\text{max}$, $R$ from $1$ to $R_\text{max}$ and  $p$ from $0$ to $p_\text{max}$. Then, the values $(K,R,p)$ which maximize BIC are chosen.

\subsubsection{Time complexity}

The proposed EM algorithm includes forward-backward procedures  \cite{BaumWelch} at the E-step to compute the joint posterior probabilities for the HMM states and the conditional distribution (the HMM likelihood) for each time series. The time complexity of the Forward-Backward procedure used at the E-Step at each EM iteration is the one of standard $R$ state HMM for univariate $n$ observation sequences of size $m$. The complexity of this step is therefore of $\cO(R^2 nm)$ per iteration. In addition, in this regression context, the calculation of the regression coefficients in the M-step of the EM algorithm requires an inversion of a $(p+1)\times (p+1)$ matrix and $n$ multiplications associated with each observation sequence of length $m$, which is done with a complexity of $\cO((p+1)^2nm)$. The proposed EM algorithm has therefore a time complexity of $\cO(I_\text{EM}  K^2 R^2 (p+1)^2 n m)$ where $I_\text{EM}$ is the number of EM iterations, $K$ being the number of clusters.

\subsection{Approximating each cluster with a single mean time series}

Once the model parameters are estimated, we derive a time series approximation from the proposed model. This approximation provides a ``mean''
 times series for each cluster which can be considered as the cluster representative or the cluster ``centroid''. Each time point of the cluster representative is computed by combining the polynomial regression components with both the estimated posterior regime probabilities $\hat{\gamma}_{ijkr}$ 
and the corresponding estimated posterior cluster probability $\hat{\tau}_{ik}$. Formally, each point of the cluster representative is given by: 
\begin{equation}
c_{kj}= \frac{\sum_{i=1}^n \hat{\tau}_{ik} \sum_{r=1}^{R} \hat{\gamma}_{ijkr}\hat{\bsbeta}^T_{kr}\bt_{j}}{\sum_{i=1}^n\hat{\tau}_{ik}}, \ (j=1,\ldots,m)
\end{equation}
 where $\hat{\bsbeta}_{k1},\ldots,\hat{\bsbeta}_{kR}$ are the polynomial regression coefficients obtained at convergence of the EM algorithm. This mean time series can be seen as a weighted empirical mean of the $n$ smoothed time series. The smoothed time series are computed as a combination between the mean polynomial regimes and their posterior probabilities. Finally, the vectorial formulation of each cluster approximation is written as
\begin{equation}
\bc_k = \frac{\sum_{i=1}^n \hat{\tau}_{ik} \sum_{r=1}^{R} \hat{\bW}_{ikr} \bX \hat{\bsbeta}_{kr}}{\sum_{i=1}^n\hat{\tau}_{ik}}.
\label{eq: cluster mean for the MixHMMR model}
\end{equation}

\section{Experimental study} 

\subsection{Experiments with simulated time series} 
\label{ssec: Experiments on simulated curves}

In this section, we study the performance of the developed MixHMMR model by comparing it the regression mixture model and the standard mixture of HMMs. We also consider two standard multidimensional data clustering algorithms: the EM for Gaussian mixtures and $K$-means algorithm. The models are evaluated in terms of clustering using  experiments conducted on synthetic time series with regime changes. 
 
\subsubsection{Evaluation criteria}
\label{sssec: evaluation criteria for the RHLP}
 
Two evaluation criteria are used in the simulations to judge performance of the proposed approach.  
The first criterion is the misclassification error rate between the true simulated partition and the estimated partition. The second criterion is the intra-cluster inertia
$\sum_{k=1}^K\sum_{i=1}^n \hat{h}_{ik}|\!|\by_i-\hat{\bc}_k|\!|^2,$
where $(\hat{h}_{ik})$ indicates the estimated cluster membership of $\by_i$ and $\hat{\bc}_k= (\hat{c}_{kj})_{j=1,\ldots,m}$  is the estimated mean series of cluster $k$. Each point of the mean series is given by:
\begin{itemize}
\item $\hat{c}_{kj}=  \hat{\bsbeta}^T_k \bt_j$ for the standard mixture of regression models,
\item $\hat{c}_{kj} = \frac{1}{\sum_{i=1}^n \hat{\tau}_{ik}}\sum_{i=1}^n \hat{\tau}_{ik} \sum_{r=1}^R \hat{\gamma}_{ijkr} y_{ij}$ for the standard mixture of HMMs,
\item $\hat{c}_{kj} = \frac{1}{\sum_{i=1}^n\hat{\tau}_{ik}} \sum_{i=1}^n \hat{\tau}_{ik} \sum_{r=1}^{R} \hat{\gamma}_{ijkr} \hat{\bsbeta}^T_{kr} \bt_j$ for the proposed model.
\end{itemize} 

\subsubsection{Simulation protocol}

\begin{figure}[!h] 
\centering 
\includegraphics[width=4.5 cm, height = 3.2cm]{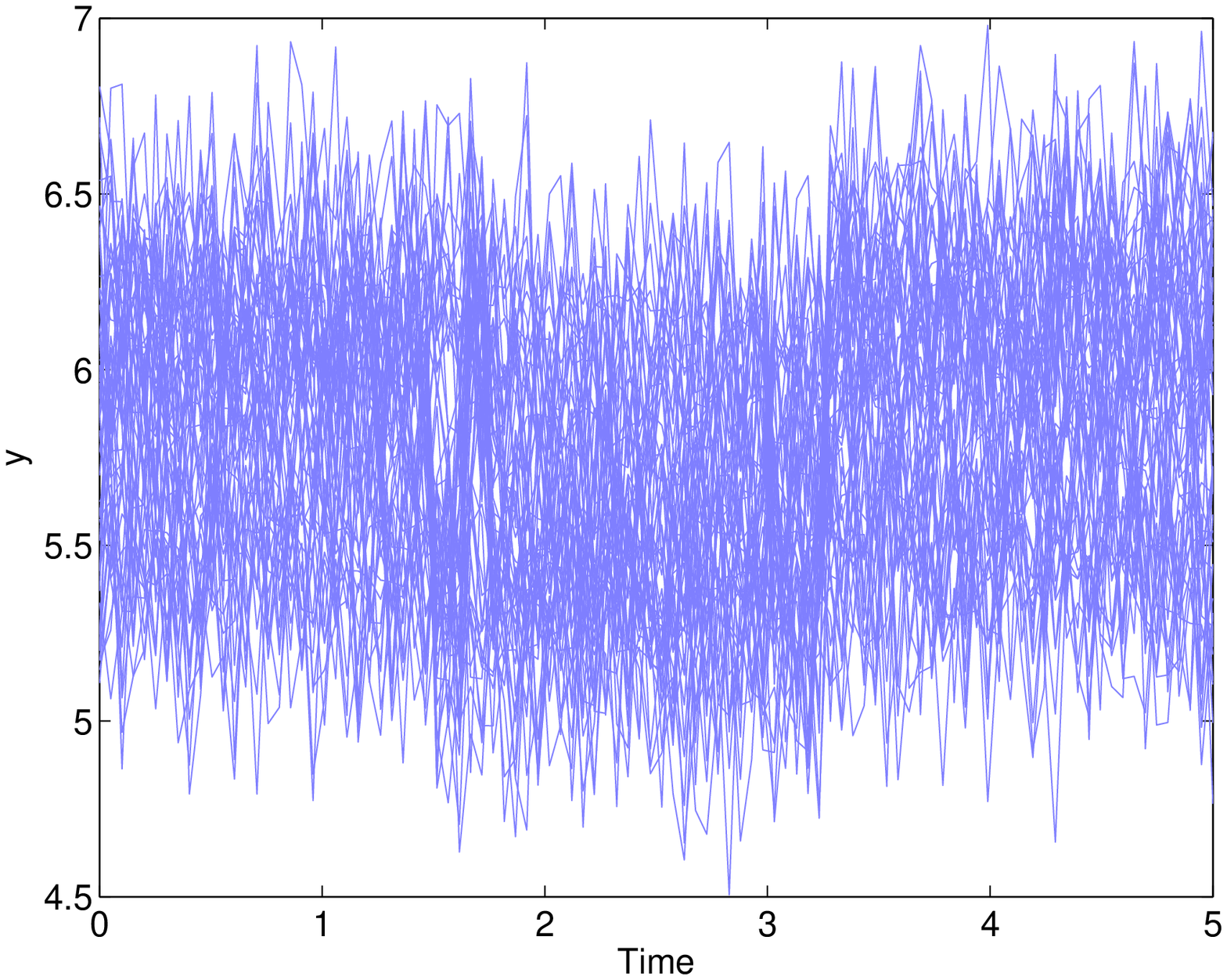}
\caption{A three-class simulated data set of $n=60$  simulated times series of size $m=100$.}
 \label{fig: example of simulated time series}
 \end{figure}

The simulated data consisted of $n$ time series of $m=100$ observations regularly sampled over the time range $[0,5]$. Each time series is generated randomly according to a particular mixture model with uniform mixing proportions ($1/K$). Each component of the mixture is a piecewise polynomial function corrupted by noise. 
The used  simulation parameters are shown in Table \ref{table: simulation parameters} and Figure \ref{fig: example of simulated time series} shows an example of simulated time series.
\begin{table}[!h]
\centering
{\small
\begin{tabular}{c llll}
\hline
Cluster & & parameters & \\
\hline
k=1& $\bsbeta_1=6.2$ & $\bsbeta_2 = 5.5$ & $\bsbeta_3 = 6$ & $\sigma=0.25$\\
k=2& $\bsbeta_1=6$ & $\bsbeta_2 = 5.3$ & $\bsbeta_3 =6.3 $ & $\sigma=0.25$\\
k=3& $\bsbeta_1=5.5$ & $\bsbeta_2 = 6$ & $\bsbeta_3 = 5.5$ & $\sigma=0.25$\\
\hline
\end{tabular}}
\caption{Simulation parameters.}
\label{table: simulation parameters}
\end{table}
\vspace*{-.5 cm}

\subsubsection{Algorithms setting}
\label{ssec: Initialization strategies and stopping rules for the (F)RHLP}

The  EM algorithm for the proposed MixHMMR model and the EM (Baum-Welch) algorithm for Hidden Markov Model Regression are initialized as follows. The parameters $\bsbeta_{kr}$ and $\sigma^2_{kr}$ for $k=1,\ldots,K$ and $r=1,\ldots,R$ are initialized from a randomly drawn partition of the time series. 
For each randomly drawn cluster $k$, we fit $R$ polynomials of coefficients $\bsbeta_{kr}$ from $R$ uniform segments of the time series of this cluster and then we deduce the value  of $\sigma^2_{kr}$. The initial HMM state probabilities are set to $\bspi=(1,0,\ldots,0)$ and the initial transition probabilities are set to $\bA_{k \ell r}=0.5$ for  $\ell \leq r \leq \ell+1$. For the  regression mixture model, the parameters $\bsbeta_k$ and $\sigma^2_{k}$ are directly estimated by fitting $R$ polynomial regression models to the randomly drawn clusters of data. 
All the EM algorithms are stopped when the relative variation of the optimized log-likelihood function between two iterations is below a predefined threshold, that is $|\frac{\cL^{(q+1)}-\cL^{(q)}}{\cL^{(q)}}| \leq 10^{-6}$  or when the iteration number reaches 1000.  We use $10$ runs of EM and the solution providing the highest log-likelihood is chosen.

\subsubsection{Obtained results}

\begin{figure*} 
\centering
\includegraphics[width=4cm]{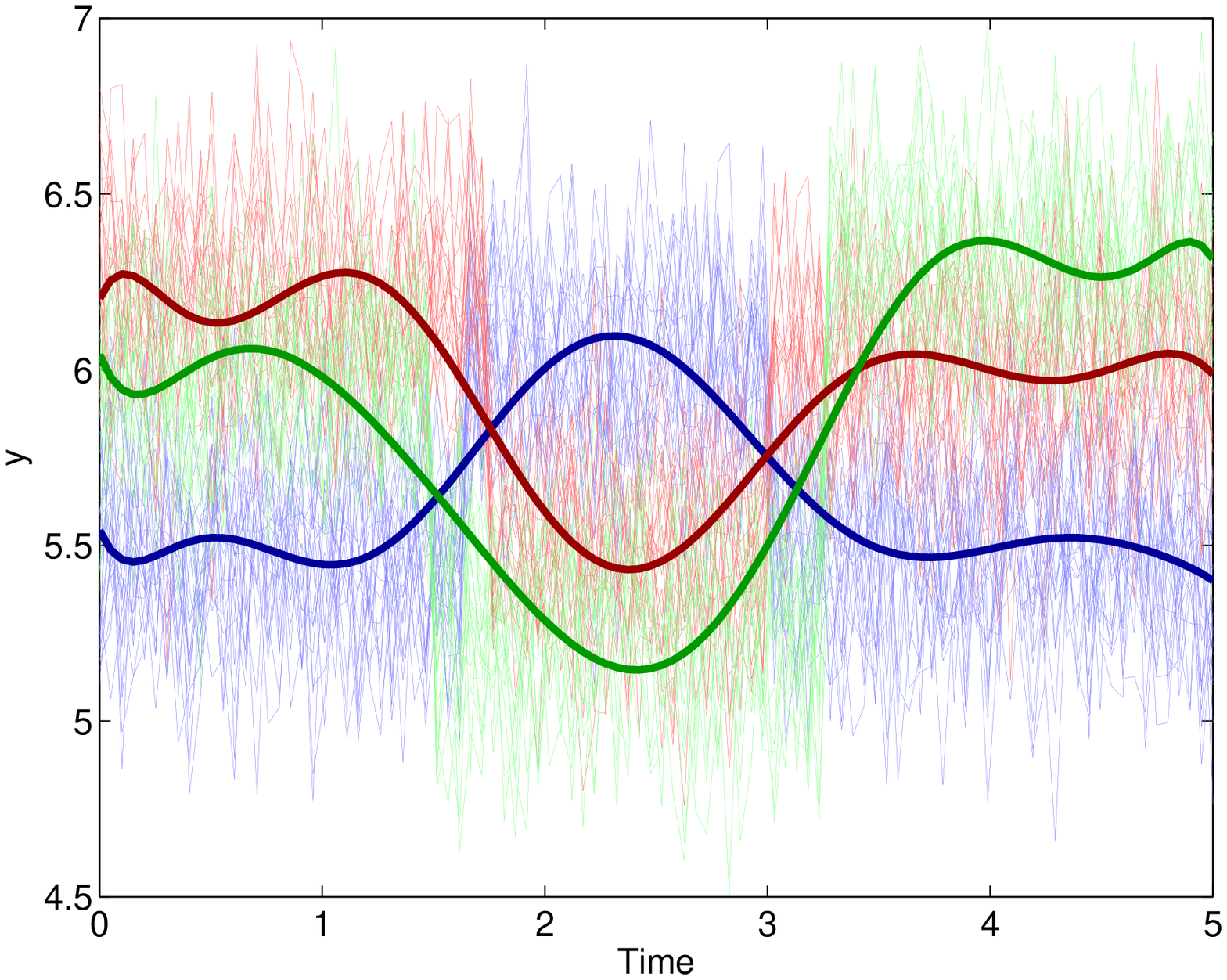}
\includegraphics[width=4cm]{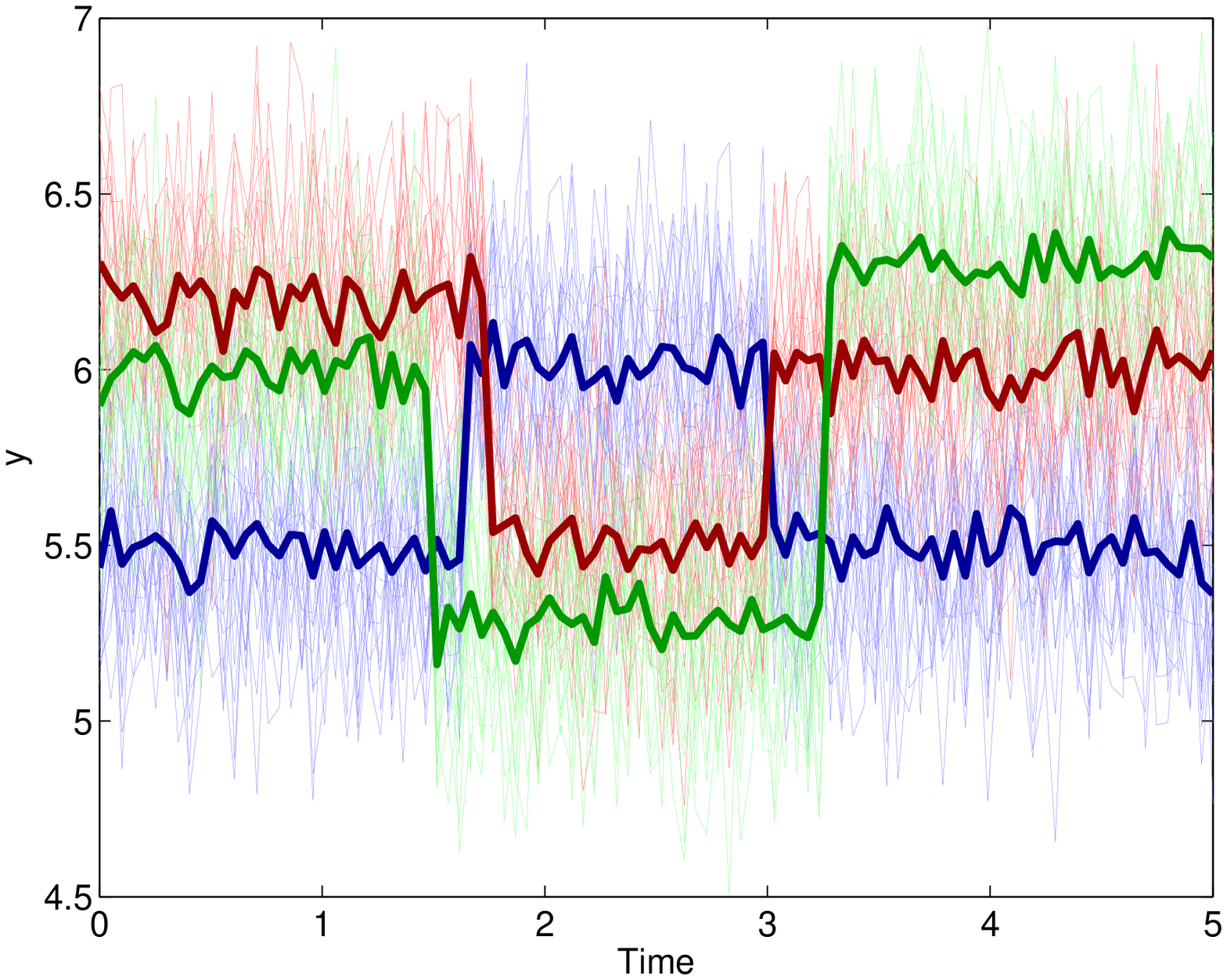}
\includegraphics[width=4cm]{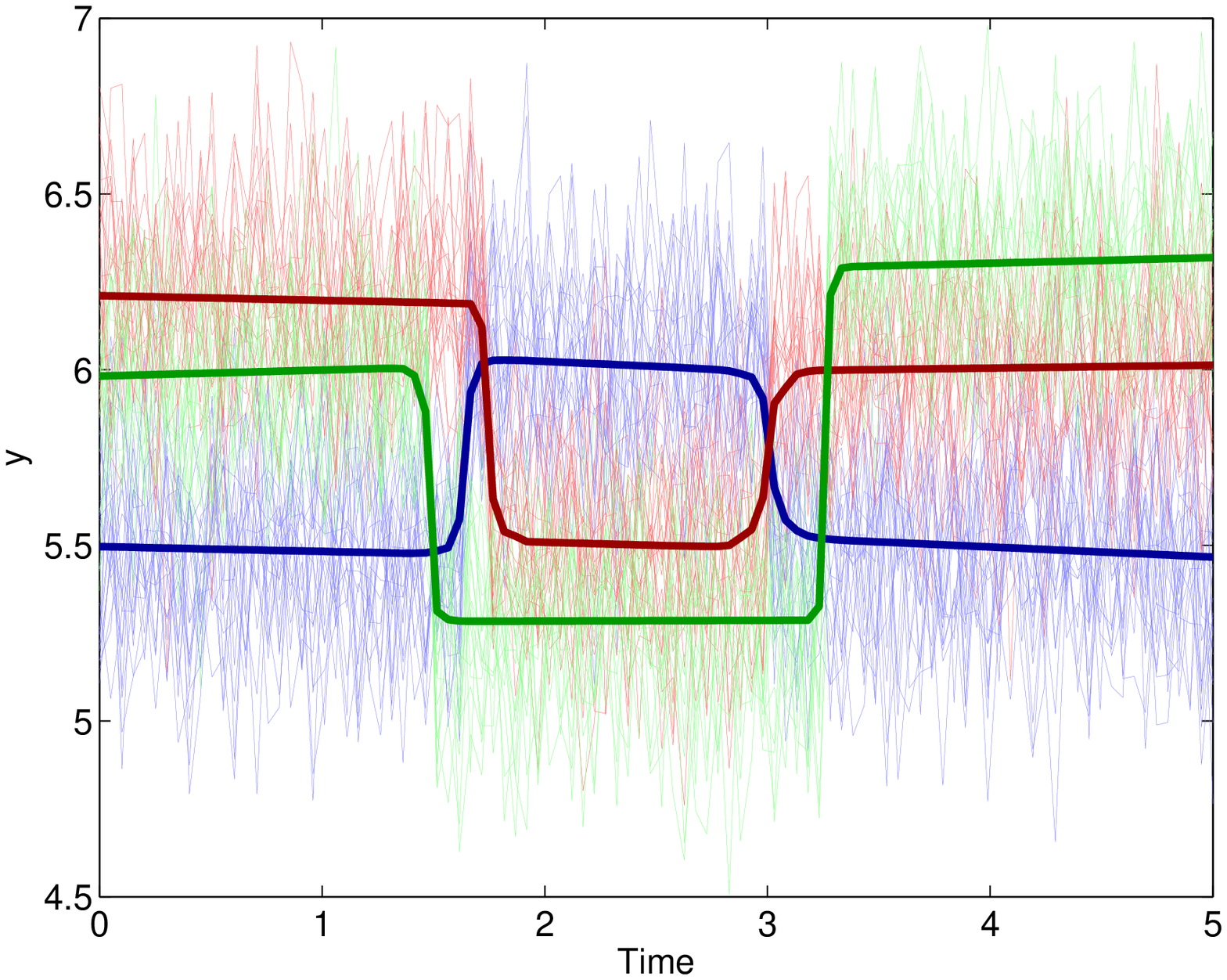}
\caption{\label{fig. estimated partitions sim data} Clustering results for the simulated time series shown in Figure \ref{fig: example of simulated time series} obtained with $(K=3,p=9)$ for the regression mixture (left), $(K=3,R=3)$ for the mixture of HMMs (middle) and  $(K=3,R=3,p=1)$ for the proposed approach (right).}
\end{figure*}

Table \ref{table. calssif and inertia results for simulations} gives the obtained misclassification error rates and the intra-cluster inertias averaged over 10 randomly drawn samples. It can be clearly observed that the proposed approach outperforms the  other approaches as it provides  more accurate classification results and small intra-class inertias.
Indeed, applying the proposed approach for clustering time series with regime changes provides accurate results, with regard to the identified clusters, as well as for approximating each set (cluster) of time series. 
This is attributed to the fact that the proposed MixHMMR model, thanks to its flexible   formulation, addresses the better both the problem of time series heterogeneities by the mixture formulation and the dynamical aspect within each homogeneous set of time series, by the underlying unobserved Markov chain. We can also observe that the standard EM for GMM  and standard $K$-means are not well suitable for this kind of longitudinal data.
\begin{table}
\centering
\small
\begin{tabular}{ lcc}
\hline
 & Misc. error rate & Intra-cluster inertia\\
\hline
Standard $K$-means	& 15 \% &  503.8434\\
Standard EM for GMM  	&  13 \% &  467.9951\\
Mixture of regressions 	& 7 \% &  495.7951\\
Mixture of HMMs 		& 6\% 	& 387.9656\\
Proposed approach		& 3 \%  & 366.2492\\
\hline
\end{tabular}
\caption{Misclassification error rates and the values of intra-cluster inertia obtained with all the algorithms.}
\label{table. calssif and inertia results for simulations}
\end{table}
 Figure \ref{fig. estimated partitions sim data}  shows partition of the time series  obtained with the three regression mixture based approaches and the corresponding cluster representatives.

%



\subsection{Clustering the real time series of switch operations} 

This section is devoted to the application of proposed clustering approach to real time series. 
\subsubsection{The used database}
\begin{figure}[h]
\centering
\includegraphics[width=5cm, height = 3.3 cm]{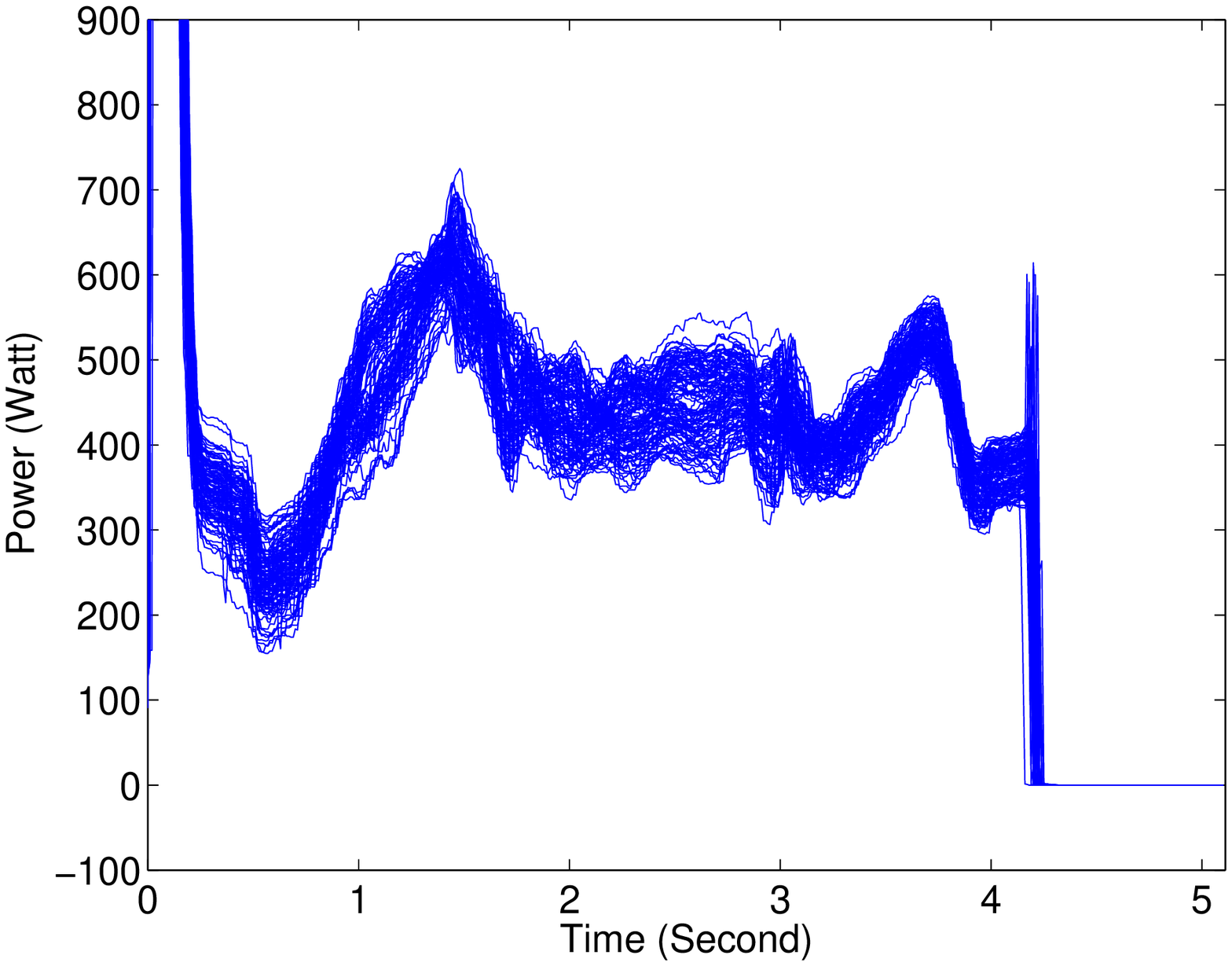}
\caption{\label{fig. railway times series}Time series of the switch operations (115 curves).}
\end{figure}

The used time series in this section are the real switch operations.  These time series present regime changes (see Figure \ref{fig. railway times series}) due to the operating process for the switch mechanism which is composed of several electromechanical movements. 

As we mentioned it in the introduction, the aim is to detect non-normal times series for a diagnosis prospective.   An important preliminary task of this diagnosis task is the automatic identification of groups of switch operations having similar characteristics. 
For this purpose, we use the proposed EM algorithm for clustering these time series.

With this diagnosis specificity, we assume that the database is composed of two clusters, one corresponding to an operating state without defect and another corresponding to and operating state with a defect, that is $K=2$. The number of regression components of the proposed algorithm was set to $R = 6$ in accordance with the number of electromechanical phases of a switch operation and the degree of the polynomial regression $p$ was set to 3 which is more appropriate for the different regimes in the time series.

\subsubsection{Obtained results}

\begin{figure} 
\centering
\begin{tabular}{cc}
\includegraphics[width=4cm,height = 2.95cm]{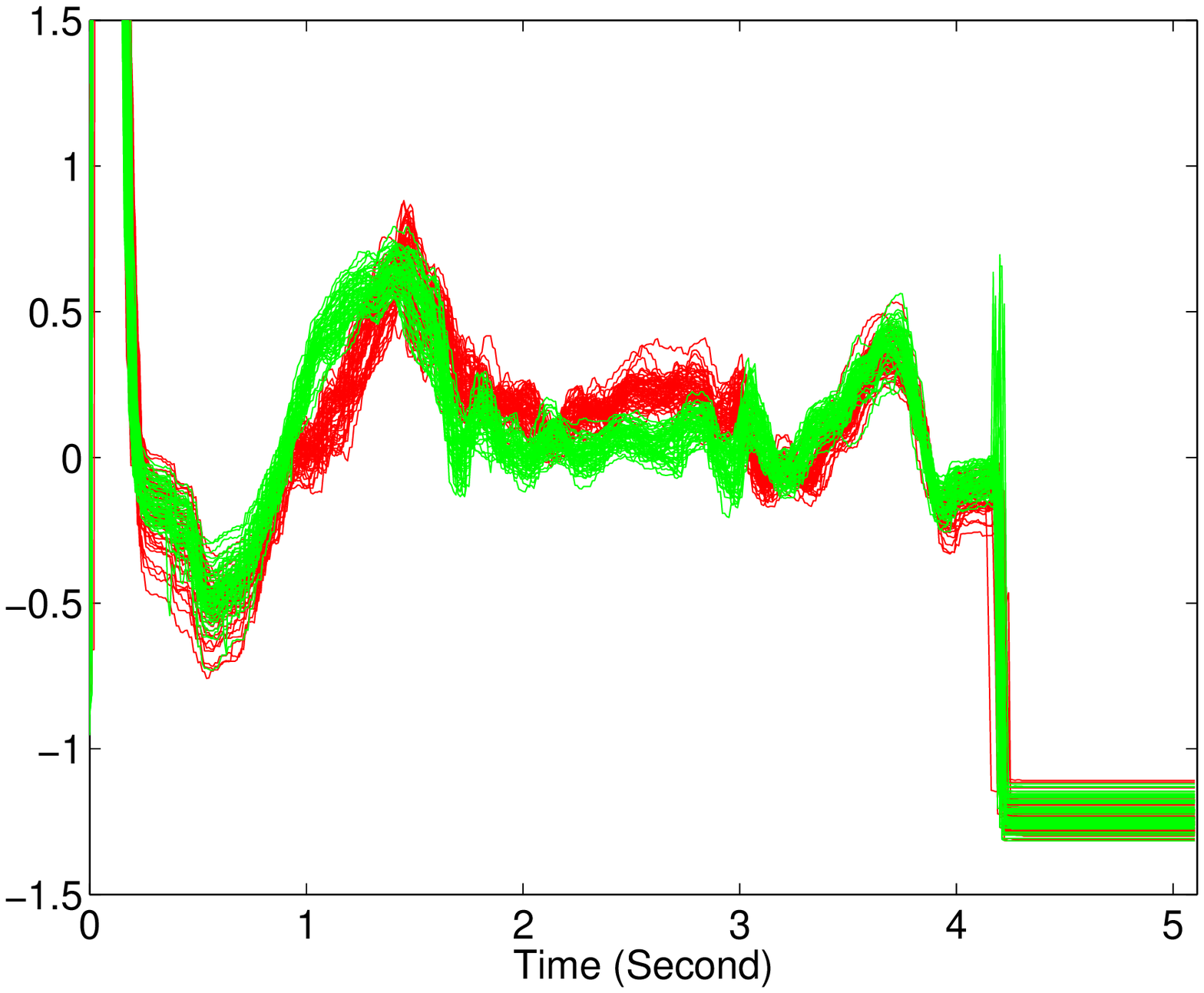}&
\includegraphics[width=3.7cm,height = 2.8cm]{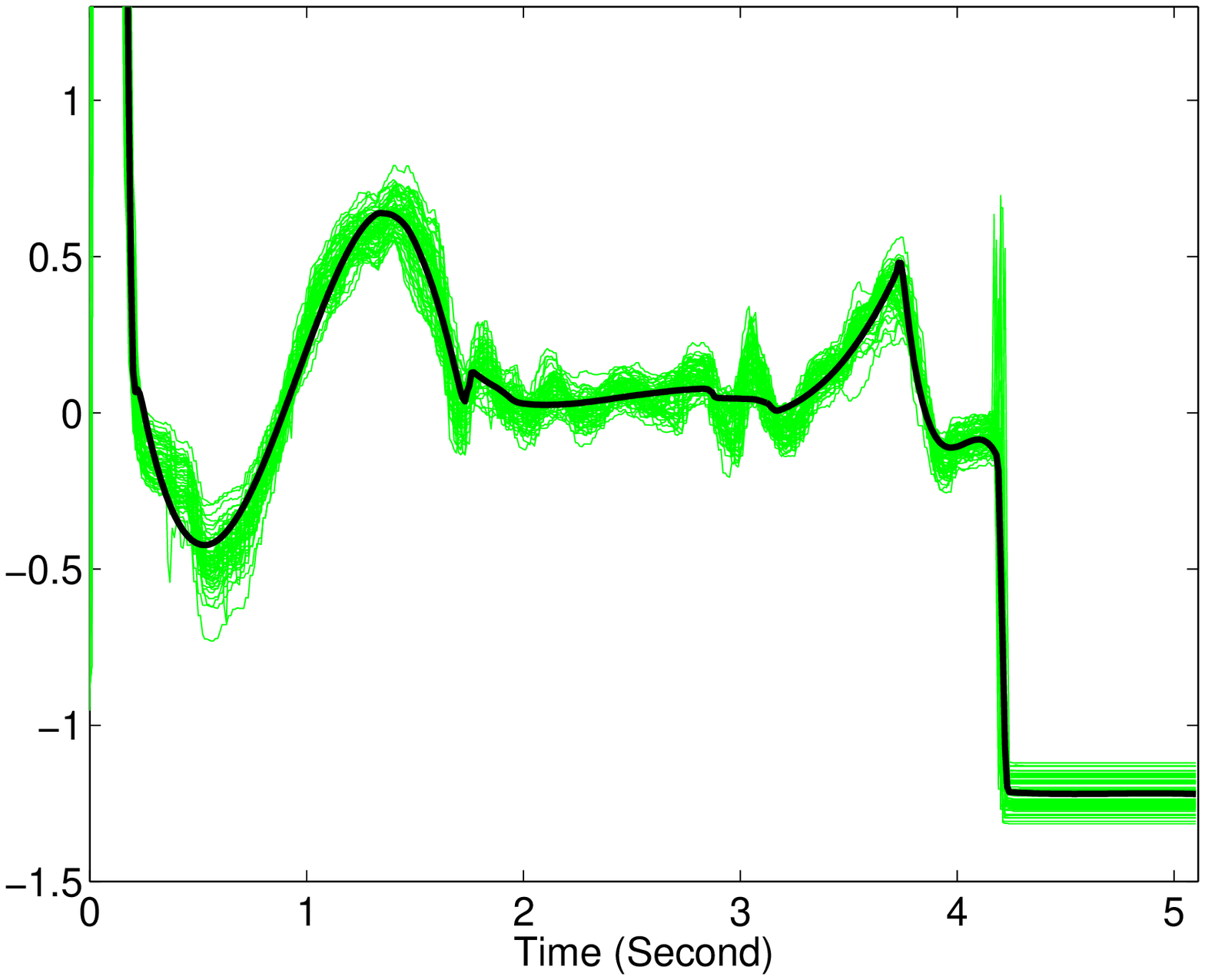}\\
\hspace*{.2 cm}\includegraphics[width=3.7cm,height = 2.8cm]{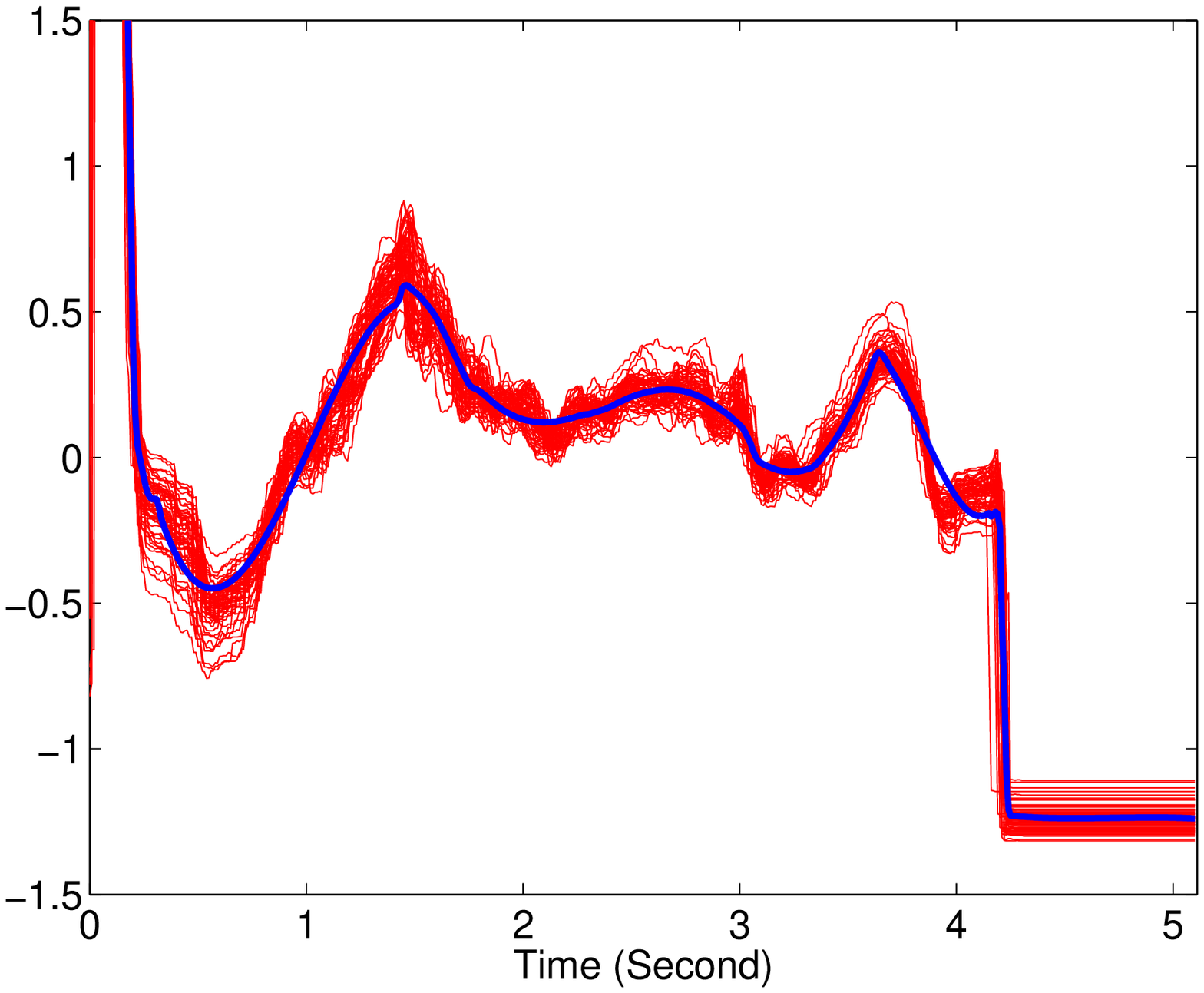}&\\
\end{tabular}
\caption{\label{fig. real time series clustering results}
Clustering results for the switch operation time series obtained for $K=6$ and $p=3$.}
\end{figure}

Figure \ref{fig. real time series clustering results} shows the graphical clustering results and the corresponding clusters approximation for the time series of the real switch operation curves.  Since the true class labels are unknown, we only consider the intra-class inertias which are given in Table \ref{table. inertia results for real data}. It can be observed on Figure \ref{fig. real time series clustering results} that the time series of the first obtained cluster (middle) and the second one (right) does not have the same characteristics  since their shapes are clearly different. Therefore they may correspond to two different stated of the switch mechanism. In particular, for the time series belonging to the first cluster (middle), it can be observed that something happened at around 4.2 Second of the switch operation. According to the experts, this can be attributed to a default in the measurement process. We note that the average running time of the EM algorithm for this experiment is about 40 S.
%
%
%
%
%
\begin{table}
\centering
{\footnotesize 
\begin{tabular}{ccccc}
\hline 
$K$-means & EM for GMM & MixReg & MixHMM   &  MixHMMR	\\ 
\hline
827.34 & 715.19 & 732.25 & 728.56 & 695.87 \\
\hline
\end{tabular}}
\caption{{\small Intra-cluster inertia for the real data.}
}
\label{table. inertia results for real data}
\vspace*{-.5 cm}
\end{table}

\section{Conclusion and future works}

In this paper, we introduced a new model-based clustering approach for time series. The proposed model consists in a mixture of polynomial regression models governed by hidden Markov chains. The underlying Markov chain allows for successively activating various polynomial regression components over time. The model is therefore particularly appropriate for clustering times series with various changes in regime. The experimental results demonstrated the benefit of the proposed approach as compared to existing alternative methods, including the regression mixture model and the standard mixture of Hidden Markov Models. At this stage, we only gave the theoretical approach for selecting a model structure trough the BIC criterion. Current experiments are concerned with this problem and future works will  discuss the problem of
 model selection.
 
\bibliographystyle{IEEEtran}
\bibliography{references}

\end{document}